\documentclass[letterpaper, 10 pt, conference]{ieeeconf}  

\IEEEoverridecommandlockouts                              

\usepackage{epsfig} 
\usepackage{amsmath} 
\usepackage{algorithm}
\usepackage{algpseudocode}
\usepackage{algorithmicx}
\usepackage{amssymb}  

\title{\LARGE \bf
Leveraging Simulation-Based Model Preconditions for Fast Action Parameter Optimization with Multiple Models
}

\author{M. Yunus Seker$^{1}$ and Oliver Kroemer$^{1}$
\thanks{*
Supported by 
NSF Grants No. CMMI-1925130 and IIS-1956163.}
\thanks{$^{1}$The Robotics Institute, Carnegie Mellon University, Pittsburgh, PA, USA 
        {\tt\small \{mseker, okroemer\}@andrew.cmu.edu}}%
}

\begin{document}

\maketitle
\thispagestyle{empty}
\pagestyle{empty}

\begin{abstract}
Optimizing robotic action parameters is a significant challenge for manipulation tasks that demand high levels of precision and generalization. Using a model-based approach, the robot must quickly reason about the outcomes of different actions using a predictive model to find a set of parameters that will have the desired effect. The model may need to capture the behaviors of rigid and deformable objects, as well as objects of various shapes and sizes. Predictive models often need to trade-off speed for prediction accuracy and generalization.   This paper proposes a framework that leverages the strengths of multiple predictive models, including analytical, learned, and simulation-based models, to enhance the efficiency and accuracy of action parameter optimization. Our approach uses Model Deviation Estimators (MDEs) to determine the most suitable predictive model for any given state-action parameters, allowing the robot to select models to make fast and precise predictions. We extend the MDE framework by not only learning sim-to-real MDEs, but also sim-to-sim MDEs. Our experiments show that these sim-to-sim MDEs provide significantly faster parameter optimization as well as a basis for efficiently learning sim-to-real MDEs through finetuning. The ease of collecting sim-to-sim training data also allows the robot to learn MDEs based directly on visual inputs and local material properties.

\end{abstract}


\section{INTRODUCTION}
Manipulation tasks often require robots to arrange objects in a precise manner while generalizing across geometric and material variations. For example, when placing food on a plate according to an example picture, the robot will need to adapt the placement parameters to the shape and rigidity of the food being arranged. Predictive models, such as simulators, allow the robot to estimate the outcomes of different action parameters across various objects. By combining such models with an optimizer, the robot can evaluate multiple action parameters and reason about their effects before executing an optimal action to achieve the desired goal in the real world. 

\begin{figure}[t]
    \centering
    \includegraphics[width=0.98\linewidth]{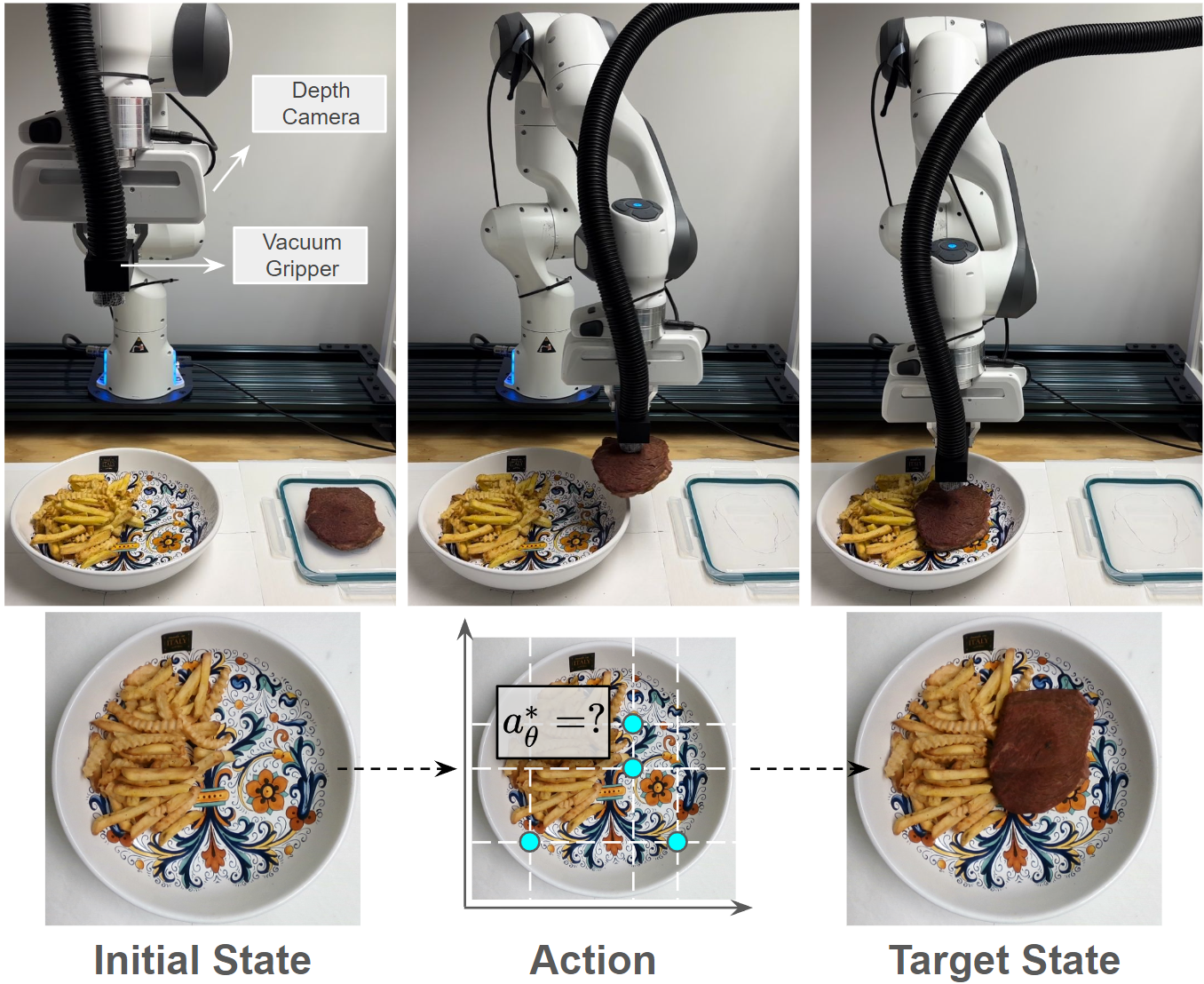}
    \caption{(Top) Illustration of the plating action. The robot is equipped with a wrist camera and a vacuum gripper. (Bottom) The robot is given initial and target scenes; the task is to find the best action to reach the target scene.}
    \label{fig:introduction}
    \vspace{-0.6cm}
\end{figure}


A significant limitation of such a model-based approach is that the quality and efficiency of the outcome rely heavily on the accuracy and speed of the predictive model. Different predictive models trade-off between speed, versatility, and accuracy. Analytical models are known for their speed in making predictions, yet they often struggle to predict complex interactions accurately, e.g., deformable objects. Simulation models provide high precision for complex scenarios, modeling rigid and deformable objects of any shape or size, but they suffer from prolonged running times. Learned models provide accuracy and swift performance; however, their reliability diminishes quickly for state distributions beyond their training data. In light of these considerations, a prudent approach for the robot is to integrate all these models and select the best model to make predictions during each step of the optimization process according to the current state and action parameters to be evaluated. 

In this paper, we propose a framework that leverages multiple predictive models by incorporating a selection mechanism that determines the most suitable model for each optimization step based on the state-action parameters. 
The robot learns a set of Model Deviation Estimators (MDEs) \cite{lagrassa2022learning} to estimate the regions of the state-action space in which each model is sufficiently accurate, i.e., its predicted values are close to the true values. We refer to these regions as the model's \emph{preconditions} as the model is unreliable outside of these regions. Using the MDEs, the optimizer evaluates a parameter setting by selecting the fastest model from the set of models whose preconditions are fulfilled. 

Previous work on MDEs \cite{lagrassa2022learning} has focused on learning deviations between model predictions and real-world outcomes. We will refer to these as sim-to-real (S2R) MDEs. We extend the MDE framework by introducing sim-to-sim (S2S) MDEs, which compare one model's predictions with that of another simulation model. In particular, we focus on learning S2S MDEs for comparing models to slow high-fidelity simulators. While S2S MDEs do not consider real-world prediction accuracy, we show that they are still a valuable tool for identifying regions in which a faster model is more suitable for evaluating parameters. The ease of collecting large training datasets for S2S MDEs also allows the robot to learn MDEs with visual height maps and material property masks as inputs. Furthermore, our experiments show that the S2S MDEs can be finetuned on limited amounts of real-world data to acquire S2R MDEs in a more sample-efficient manner. The contributions of this paper can thus be summarized as:

\begin{itemize}
\item The utilization of multiple predictive models to speed up the optimization of robot action parameters.
\item Sim-to-sim MDEs for speeding up the prediction processing using only simulation-based training.
\item Creating vision-based MDEs for selecting fast and accurate predictive models based on heightmap state representations, rather than low-dimensional state vectors.
\item Incorporating material property masks as MDE inputs to handle both rigid and deformable object.
\item Demonstrating the sample-efficient transition of our  MDE framework from simulation-based training to real-world predictions through finetuning.
\end{itemize}

\section{RELATED WORK}
Model and skill preconditions have been studied in various works to detect model uncertainties for accurate action planning, recovery, and deformable object manipulation \cite{lwtunreliablemodels, lwtdynamicmodel,sharma2020relational, kroemerspatialskillpreconditions}. More recently, MDEs \cite{lagrassa2022learning} have been proposed as powerful model precondition learning frameworks that allow planning with multiple models with changing fidelity. MDEs have also been shown to be successful in adapting dynamics models to new environments by focusing on similar regions of the model with the real-world \cite{mitrano2023focused}. Similarly, \cite{lagrassa2024taskoriented} used MDEs for actively learning model preconditions in tasks where real-world data collection is costly or risky, focusing on areas where a dynamics model is accurate for planning. 
In this study, we further extend MDEs by incorporating visual and material property masks and employing a data-efficient S2S approach. 

Using multiple models to speed up planning tasks has also been investigated in various robotic applications, such as planning actions for movable objects \cite{saleem2020planning} and underwater planning an I-AUV \cite{8593604}. While providing efficient planning, these models do not require an advanced model selection. In our work, we utilize multiple models with advanced MDEs capable of processing high-dimensional visual data.

Integrating material properties into robotic applications is also a commonly studied topic for various planning and manipulation tasks, such as estimating the friction coefficient of objects \cite{friction} and learning to slide objects \cite{slide}. These studies mainly focus on predicting the material properties of the objects based on real-world observations \cite{seker2023estimating, matl2020inferring, pmlr-v87-clarke18a}. By contrast, our approach focuses on incorporating material properties into the decision-making process to allow better context-aware optimization. 

Our optimization framework is designed to find the optimal action parameters to reach a target state. Previous research in this area primarily focuses on System Identification \cite{sanchezgonzalez2018graph, li2019learning} and differentiable simulations \cite{jatavallabhula2021gradsim, drake} to predict optimal actions based on real-world observations \cite{pmlr-v205-antonova23a, ma2022risp}. While existing works mainly employ a single simulator to find optimal actions, our approach utilizes a set of multiple models during the optimization to enable fast and accurate predictions.

\begin{figure}[t]
    \centering
    \includegraphics[width=\linewidth]{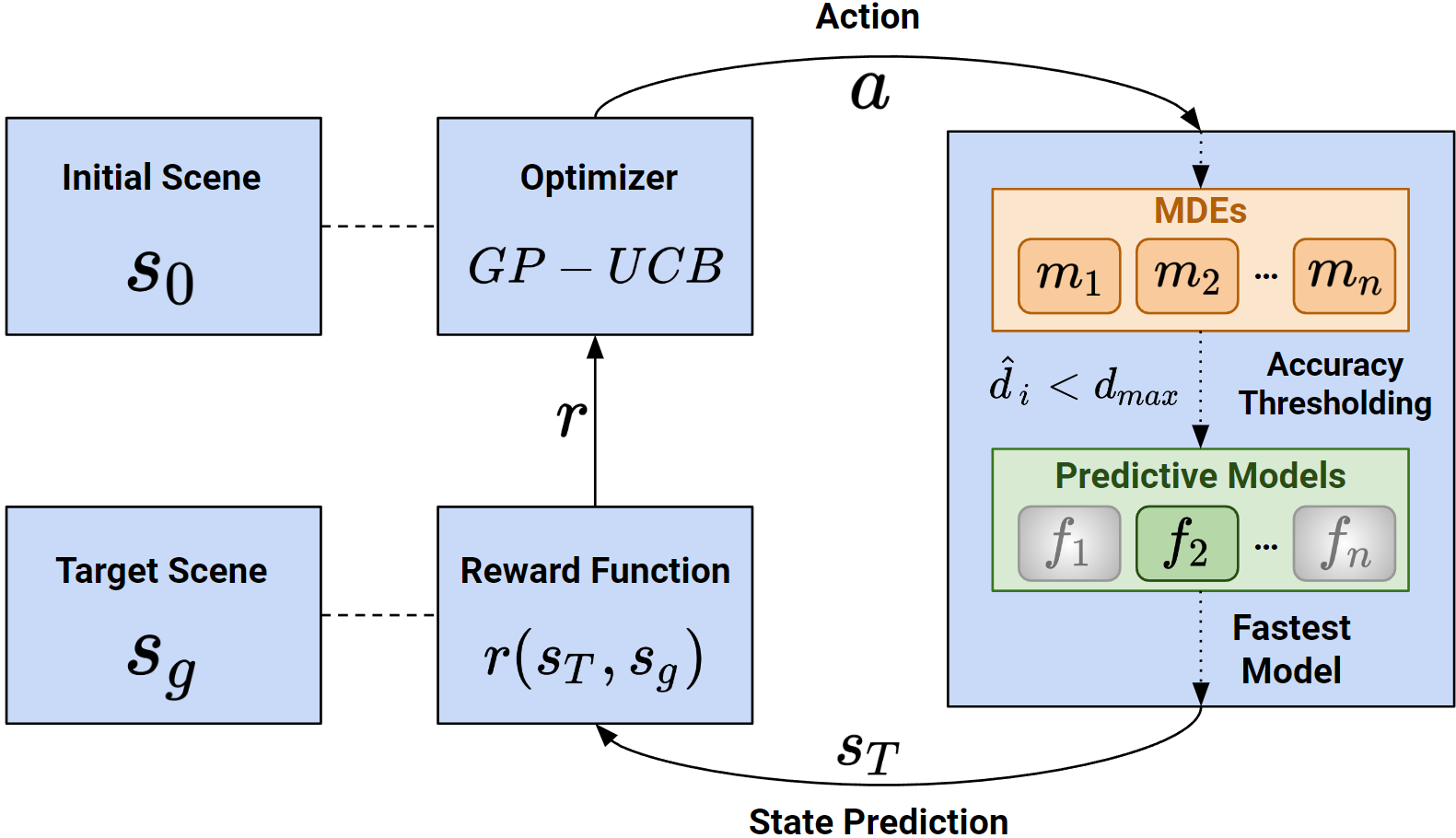}
    \caption{Overview of our framework. Given an initial state, the robot employs an optimizer guided by a reward function to predict the optimal action to achieve a target state. Throughout this optimization process, the robot leverages MDEs to dynamically select the most appropriate model from a family of predictive models.}
    \label{fig:overview}
    \vspace{-0.6cm}
\end{figure}
\section{TECHNICAL APPROACH}
In this section, we present the optimization framework, the multiple predictive models, and the model selection process using MDEs. We also introduce sim-to-sim MDEs, how they can be finetuned, and we propose an architecture for incorporating image data as input to generalize across object shapes and material properties.

\subsection{Optimization Problem Formulation} We model the robot's action selection task as a one-step goal-based reinforcement learning problem. We assume that the robot is provided with an initial state $s_{0}$ as well as a target state $s_g$. The robot must select an action $a$, which will result in a transition from the initial state $s_{0}$ to a final state $s_{T}$. The robot then receives a reward based on the similarity of the final state and the goal state $r(s_T,s_g)$. It should be noted that the target state $s_g$ may not actually be reachable from the initial state $s_0$.  For example, the target may demonstrate the task for a different object, which the robot will try to match as closely as possible with the given objects. The goal of the robot is to maximize the reward. The action $a$ is conceptualized as a collection of $m$ parameters $a = \{\theta_1, ...\theta_m\}$. 

As a model-based approach, the robot uses a predictive model $f(s_0,a)$ to estimate the final state $s_T$ for different action parameters $a$. The robot thus selects the optimal action to execute according to $a^*=\arg\max_a r(f(s_0,a),s_g)$. We perform this optimization using a GP-UCB approach \cite{Srinivas_2012, bo}. We selected a Bayesian optimization approach to encourage a global search of the action space, which will involve evaluating the prediction model $f(s_0,a)$, the focus of this paper, across a wider range of parameter values than in a local search. In practice, the robot will need to find suitable action parameters as quickly as possible to avoid delaying the overall task.

\subsection{Modeling Object Arrangement Tasks}
Our evaluations focus on optimizing actions for an object arrangement task: placing food on a plate. The initial state consists of two heightmaps $s_0 = \{I_0, I_{obj}\}$ where $I_0$ and $I_{obj}$ are the heightmap representations of the initial scene and the manipulated object respectively. The target state, $s_g = \{I_g\}$, consists of a single heightmap. We define the 2D action $a=(x, y)$ as the planar x and y coordinates at which the grasped object will be placed. For our reward function, we used the negative differences between the predicted and target heightmaps, $r(f(s_0, a), s_g) = - ||\hat I - I_g||$.

Material properties play a critical role in food manipulation tasks. The properties can be estimated using interactive perception or through observations of the objects interacting (\cite{seker2023estimating}). We assume that the material properties of the objects, such as the stiffness and density coefficients, are already known. To capture these properties, we provide the robot with a set of material property masks $\phi$. We use three channels to capture the object mass, Young's modulus, and Poisson values. These masks have the same dimensions as the heightmap and correspond to the properties of the top items at the corresponding heightmap locations. We do not include the properties of any potential hidden objects with different properties, nor do our evaluations include such confounder objects.

\subsection{Individual Predictive Models for Object Arrangement}

We assume that the robot has access to a family of predictive models, $F=\{f_1, f_2, ...,f_n\}$, consisting of $n$ different analytical \cite{GOYAL1991307,pan2020decision}, learned \cite{hafner2019learning,nagabandi2019deep, tekden2020belief}, and simulation models \cite{coumans2021,isaacgym}. Each predictive model takes the initial state and the action parameters and outputs a prediction for the resulting scene, $f_i(s_{0}, a) = s_T$. The accuracy of each model will vary across the state-action space. We assume that the predictive models in $F$ are ordered by their index, $i$, with respect to their computational expansiveness. A predictive model $f_j$ is faster than $f_k$ if $j<k$.

For the object arrangement task, we consider a heuristic model $f_1$, a learned model $f_2$, and a simulator $f_3$. The following describes each of the models in the set $F$.

\begin{figure}[t]
    \centering
    \includegraphics[width=\linewidth]{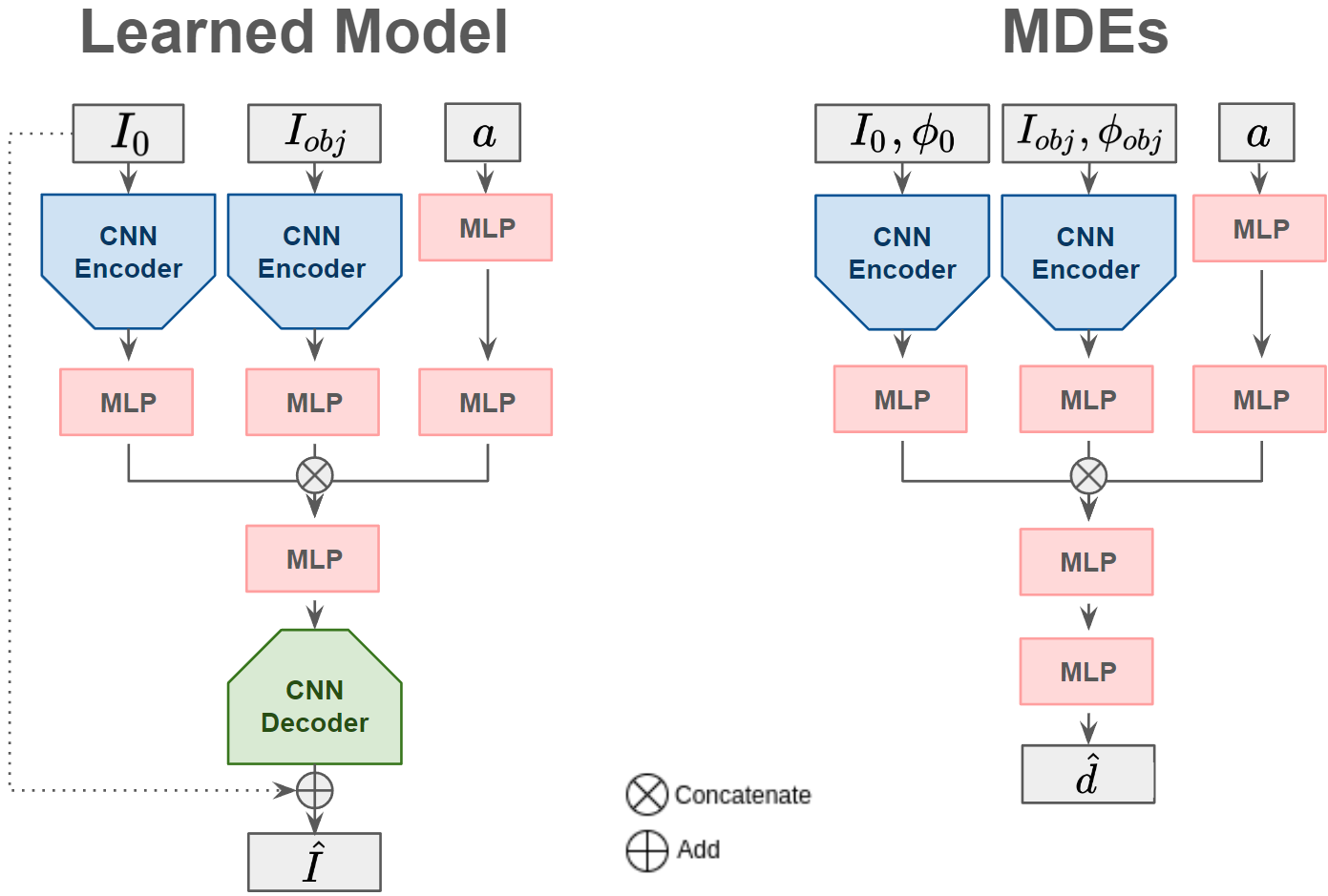}
    \caption{(Left) Neural network architecture of the learned model. (Right) Neural network architecture of the MDE models.}
    \label{fig:nn_mde}
    \vspace{-0.1cm}
\end{figure}


\textbf{\textit{Heuristic Model ($f_1$):}} For our analytical model, we used a deformable heightmap heuristic that adds the target object heightmap on top of the initial scene according to given placing action parameters. Algorithm \ref{alg:1} shows the procedure for applying the deformable image heuristics given the initial scene, the object, and the action parameters. First, the object heightmap is shifted on the x and y axes according to the given action parameters. Then, it is added on top of the initial heightmap of the scene to predict the result of the placing action. This model is computationally very efficient, but it may not be accurate for cases involving complex surface geometries or when the target object and the scene interact in a non-linear manner. Nevertheless, this model produces plausible predictions for scenarios where the target object is placed on empty and flat areas of the plate.

\begin{algorithm}[b]
\caption{Heuristic (Analytical) Model}
\begin{algorithmic}[1]
\Procedure{HeuristicModel}{$scene, obj, action$}
    \State $obj \gets \Call{ShiftImage}{obj, action}$
    \State $heuristicPrediction \gets scene + obj$
    \State \Return $heuristicPrediction$
\EndProcedure
\end{algorithmic}
\label{alg:1}
\end{algorithm}

\begin{figure}[t]
    \centering
    \includegraphics[width=\linewidth]{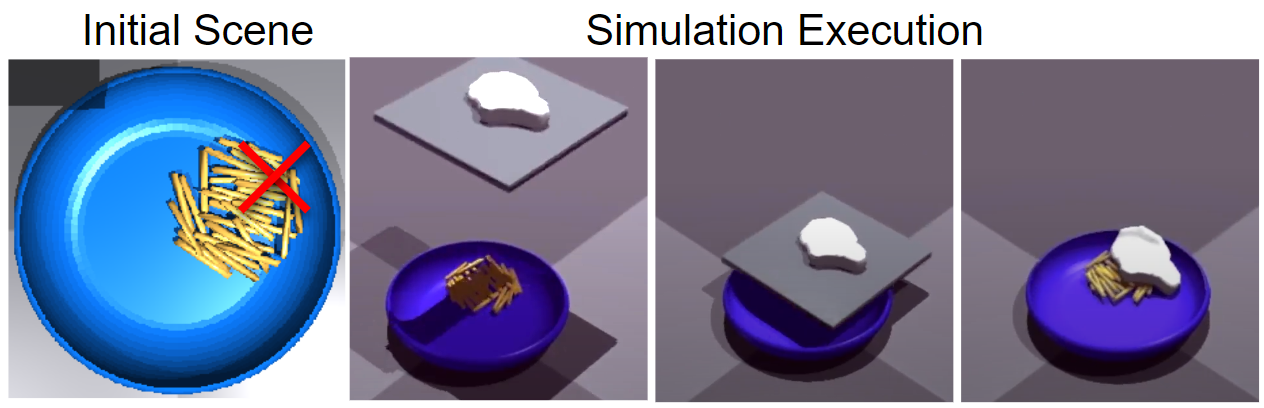}
    \caption{Isaac Gym Simulation model. The plate and fries are initialized at the start of the simulation. The steak is placed at the target location, marked by the red X, using a tray that only contacts the steak.}
    \label{fig:simulation}
    \vspace{-0.5cm}
\end{figure}
\textbf{\textit{Learned Model ($f_2$):}} For our learned model, we used a residual encoder-decoder network that predicts the final heightmap of the environment given the action parameters and the heightmaps of the initial scene and the target object. Figure \ref{fig:nn_mde} shows the neural network architecture we used for our learned model. We opted for a residual prediction model for two reasons: First, as our analytical model $f_0$ is designed based on a residual summation operation, we also wanted to design the same behavior for our learned model. Second, we wanted our model to focus on the effects of the placing action with the deformable steak instead of predicting the whole heightmap from scratch. We trained our learned model with the observations collected from the simulation shown in Figure \ref{fig:simulation} by generating a dataset with 1000 randomly sampled french fries and steak placement locations. The learned model is trained with the MSE loss between the predicted heightmaps and the ground truth observed from the simulator.

\textbf{\textit{Simulation Model ($f_3$):}} For our simulation model, we chose NVIDIA Isacc Gym \cite{isaacgym} as it provides stable deformable object interactions. Figure \ref{fig:simulation} shows an example execution of a plating action. For a given initial scene and action parameters, the plate and fries are initialized accordingly, and the steak is placed in the given location by a decreasing tray that does not collide with the fries and plate. In order to simulate the scenarios from the real world, we first used the real-to-sim pipeline detailed in \cite{seker2023estimating} in order to transfer the steak object from the real world into the simulation and initialized the fries' location based on the real-world positioning using SAM \cite{kirillov2023segment} image segmentation.
\begin{figure*}[t]
    \centering
    \includegraphics[width=\linewidth]{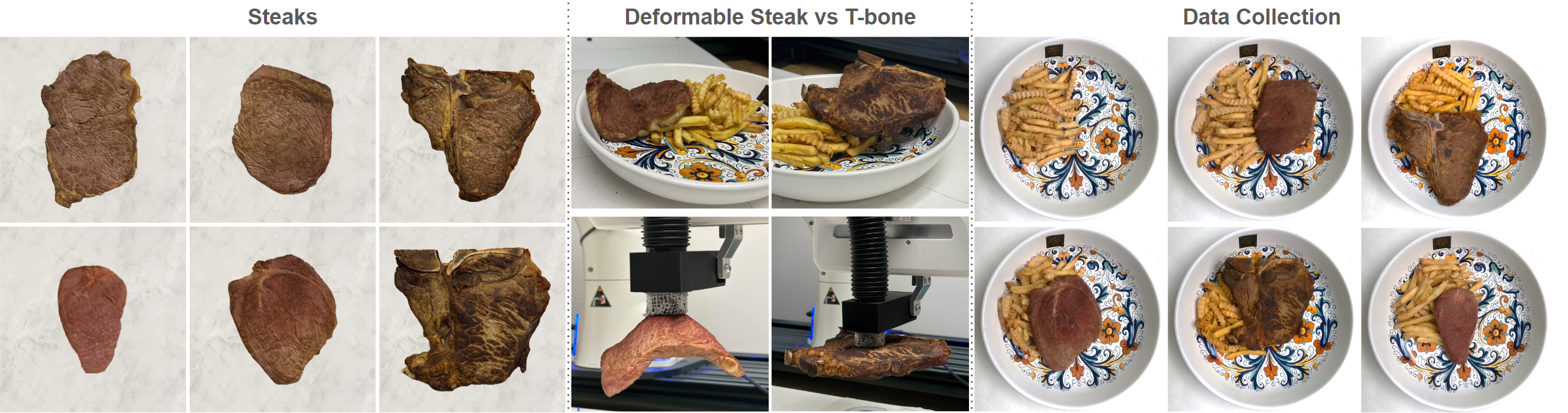}
    \caption{(Left) Four steaks are cooked rare to allow deformable behavior. Two T-bone steaks are cooked well-done to form rigid behavior. (Middle) Deformable and rigid steak comparison on the plate and on the robot vacuum gripper. (Right) Example real-world initial and target scenes.}
    \label{fig:real_world}
    \vspace{-0.5cm}
\end{figure*}

\subsection{Combining Models for Fast and Accurate Optimization}
Using the multiple models in $F$, the robot aims to \emph{quickly} optimize the action parameters by minimizing the error between the predicted final state $s_T$ and the target state $s_g$. To achieve this goal, we propose combining the multiple models to achieve accurate and fast optimization. In particular, the robot must select a model from the set $F$ for each step of the optimization process. The selected model should be as fast as possible while also being accurate. Thus, while computationally expensive simulations may generally be the most accurate and versatile of the models, other faster models should be favored if they are sufficiently precise and accurate. 


Model deviation estimators (MDEs) \cite{lagrassa2022learning} allow a robot to estimate the magnitude of a model's prediction error for different state-action parameters. By only using predictive models in scenarios where this estimated deviation is low, the robot can avoid using poor predictions in its optimization process. We refer to these accurate-prediction regions as the model preconditions, as the model can only be used reliably in these regions.

To train a sim-to-real (S2R) MDE $m_i(s,a)$ for model $f_i$, we provide the robot with a deviation function $d(s_T, f_i(s_0,a))$ for computing the deviation between the predicted state $f_i(s_0,a)$ and the actual ground truth final state  $s_T$ collected from the real system. For the object arrangement task, we define the deviation function as the L1 difference $d(s_T, f(s_0, a)) = ||\hat I - I_T||$ between the model's predicted heightmap $\hat I$ and the ground truth final heightmap $I_T$.
The MDE regressor is trained to directly predict this deviation using an L1 loss:
$$ \mathcal{L}=||m_i(s_0,a) - d(s_T, f_i(s_0,a)) ||$$ To train the S2R MDE, the robot requires samples of both accurate and inaccurate predictions from a suitably wide range of the state-action parameter space. Recent works have explored using active learning approaches to make the costly real-world data collection process more sample efficient \cite{lagrassa2024taskoriented}.

During each step of the optimization process, the robot must select the best model for the current state-action parameters and MDE outputs. A predictive model $f_i$ is considered to be applicable, i.e., within its preconditions, if its predicted deviation is less than a predefined threshold, $ m_i(s_0, a) < d_{max}$. In our experiments, the hyperparameter $d_{max}$ is set to $0.4$. Excluding the models whose preconditions are not fulfilled, the robot selects the fastest remaining model to estimate the next step $s_T$ for the current action parameter $a$. In this paper, we assume that the simulation model $f_n$ is always within its preconditions, and thus, there is always a model that can be queried. 

After selecting a model, the robot subsequently computes the reward function $r$ based on this prediction and continues its search for the optimal $a^*$ according to the outcomes of the different action trials using GP-UCB. We run the optimization for a fixed number of 50 steps before selecting the $a^*$ that will be executed on the real robot. 
It is important to note that the best model to use may change in each step of the optimization as the robot considers different state-action regions.

\subsection{MDE Architecture for Object Arrangement Tasks}

Previous approaches to learning MDEs have focused on using low-dimensional state representations, e.g., object positions. However, generalizing over different object geometries and material properties often requires a more versatile input representation. 
Our MDE architecture therefore takes in image inputs to capture both the scene geometry and the corresponding material properties. 

As shown in Fig. \ref{fig:nn_mde} right, the MDE encodes the inital scene and object information separately using CNN encoders. The corresponding height map $I$ and material mask $\phi$ images are combined as separate channels for the input image. This early fusion approach allows the robot to reason about the geometry and material properties at the pixel level. The resulting embeddings are then concatenated together with the action embedding. The final fully connected layers then regress the estimated deviation value.

The proposed architecture does not segment out individual objects, i.e., two potatoes next to each other may appear identical to one large potato. However, segmentations can be noisy especially for heaped objects in a scene, and the object being placed is always a single object in our experiments. Segmenting objects was therefore not necessary for our task.

\subsection{S2S MDEs for Fast Optimization and Efficient Transfer}

The Sim-to-Real MDEs allow the robot to select faster models when they provide sufficiently accurate effect predictions. However, for speeding up predictions, this constraint can be unnecessarily restrictive. Instead, if two models provide similar predictions, then the faster one should still be used. We therefore propose using Sim-to-Sim (S2S) MDEs to speed up the parameter optimization process. 

The structure and training of the S2S MDEs is still largely the same as for S2R MDEs. However, rather than using ground truth final states $s_T$ for learning the MDEs, we instead use the predicted state of another model. The training loss for MDE $m_{ij}$ between models $f_i$ and $f_j$ thus becomes
$$ \mathcal{L}=||m_{ij}(s_0,a) - d(f_i(s_0,a), f_j(s_0,a)) ||$$

The data collection process is however a lot easier for S2S MDEs as their is no need for real world data collection. The robot can simply use both models to simulate the outcomes for a wide range of object shapes, materials, and poses as well as action parameters. The resulting set of diverse samples allow the robot to learn a precise MDE model across a wide range of parameter values. In our experiments, the S2S MDEs are trained on 1000 samples.

For our optimization process, we train S2S MDEs between each of the models and the high-fidelity simulator. As previously noted, the simulator is always treated as being within its preconditions to represent a fallback option. The robot then uses the S2S MDEs rather than the S2R MDEs to select which other models to consider. The robot chooses the fastest model, as before, from the resulting set.

\subsection{Fine-tuning S2S to S2R MDEs with Real-world Data}

S2S MDEs provide a means of directly speeding up the optimization process without requiring any real-world data collection. However, in practice, S2R MDEs will ultimately provide more accurate predictions as they take into consideration the sim-to-real gap. This is especially true in cases when a faster model is more accurate than the simulation model, e.g., due to simulation instability. 

To learn S2R MDEs, we also explore fine-tuning the S2S MDEs using real-world data, with the S2S training serving as a pretraining. To fine-tune our S2S MDEs, we freeze the weights of the S2S networks except the last layer and train our MDE models with real-world data to develop S2R MDEs. Our experiments showed that S2S MDEs can be quickly adapted to S2R MDEs by fine-tuning with a few real-world observations. This adaptation capability allows data-efficient training of S2R MDEs and creates a significant advantage compared to the systems that rely completely on costly real-world data collection.

\begin{table}[b]
\centering
\caption{Evaluation of Predictive Models}
\resizebox{\linewidth}{!}{
\begin{tabular}{c|c|c|c}
 &
  Heuristic Model &
  Learned Model &
  Simulation Model \\ \hline
\begin{tabular}[c]{@{}c@{}}Isaac Gym Dataset\\  Heightmap RMSE ($cm^2$)\end{tabular} &
  \begin{tabular}[c]{@{}c@{}}1.7558\\ (0.5474)\end{tabular} &
  \begin{tabular}[c]{@{}c@{}}0.4639\\ (0.1915)\end{tabular} &
  \begin{tabular}[c]{@{}c@{}}0.023\\ (0.007)\end{tabular} \\ \hline
\begin{tabular}[c]{@{}c@{}}Real-World  Dataset\\ Heightmap RMSE ($cm^2$)\end{tabular} &
  \begin{tabular}[c]{@{}c@{}}5.6945\\ (2.8731)\end{tabular} &
  \begin{tabular}[c]{@{}c@{}}7.1689\\ (0.0303)\end{tabular} &
  \begin{tabular}[c]{@{}c@{}}1.4041\\ (1.6889)\end{tabular} \\ \hline
\end{tabular}
}
\label{tab:model_accuracy}
\end{table}

\begin{table}[b]
\centering
\caption{Evaluation of MDE Models}
\resizebox{\linewidth}{!}{%
\begin{tabular}{c|c|c}
Outputs scaled [0-1] &
  Heuristic MDE &
  Learned MDE \\ \hline
\begin{tabular}[c]{@{}c@{}}S2S MDEs on \\ Isaac Gym Dataset\end{tabular} &
  \begin{tabular}[c]{@{}c@{}}0.1359 \\ (0.029)\end{tabular} &
  \begin{tabular}[c]{@{}c@{}}0.0876\\ (0.025)\end{tabular} \\ \hline
\begin{tabular}[c]{@{}c@{}}S2S MDEs on \\ Real-world Dataset\end{tabular} &
  \begin{tabular}[c]{@{}c@{}}0.1949\\  (0.224)\end{tabular} &
  \begin{tabular}[c]{@{}c@{}}0.2735 \\ (0.056)\end{tabular} \\ \hline
\begin{tabular}[c]{@{}c@{}}Finetuned S2R MDEs on \\ Real-world Data\end{tabular} &
  \begin{tabular}[c]{@{}c@{}}0.1490 \\ (0.019)\end{tabular} &
  \begin{tabular}[c]{@{}c@{}}0.1472 \\ (0.027)\end{tabular} \\ \hline
\end{tabular}
}
\label{tab:mde_accuracy}
\end{table}


\begin{figure*}[t]
    \centering
    \includegraphics[width=\linewidth]{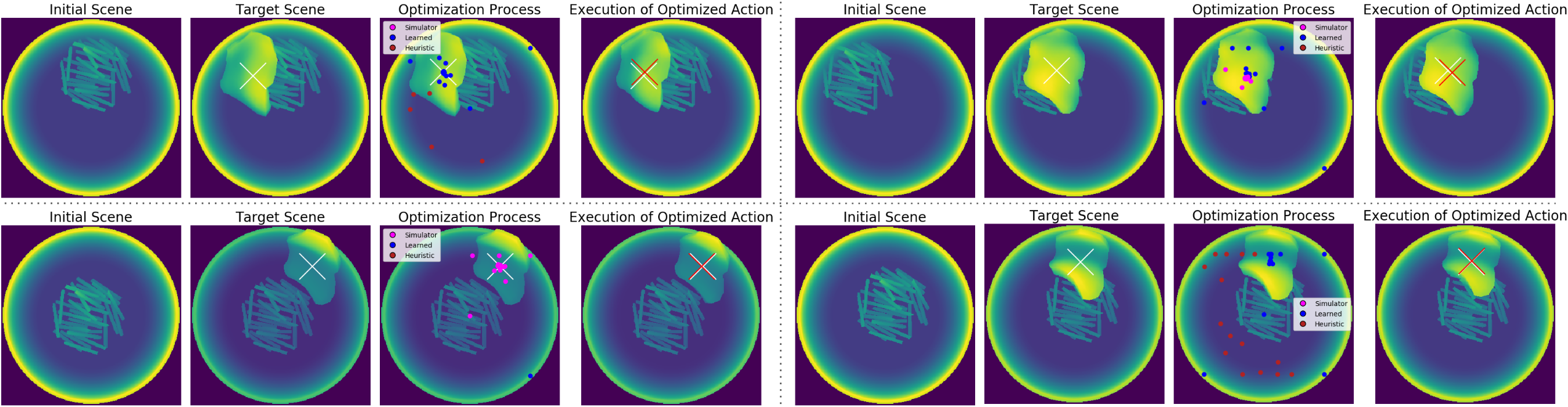}
    \caption{Example results for optimizing the plating action based on given initial and target scenes from Isaac Gym Dataset. 
    }
    \label{fig:sim_results}
    \vspace{-0.5cm}
\end{figure*}
\begin{figure}[b]
    \centering\vspace{-5mm}
    \includegraphics[width=\linewidth]{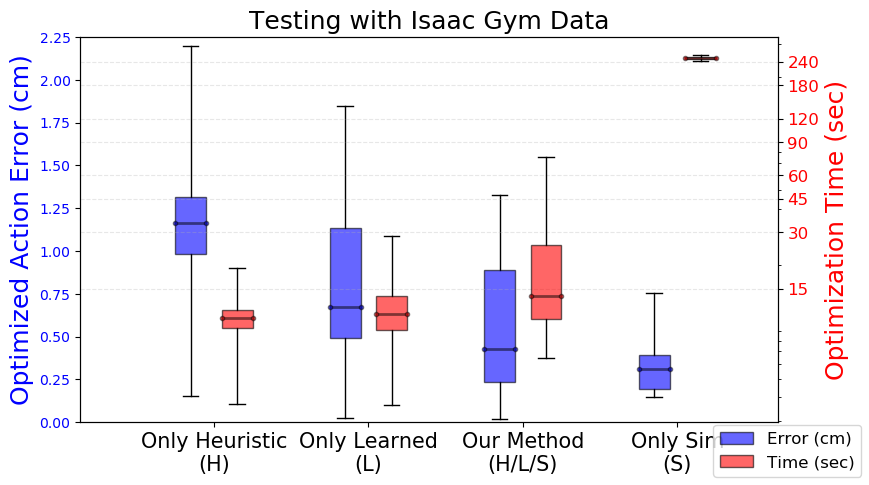}\vspace{-3mm}
    \caption{Comparison of our model with the baselines for Isaac Gym dataset. (left axis) optimized action  (right axis, log scale) optimization time.}
    \label{fig:test_sim}
\end{figure}
\begin{figure}[b]
    \centering
\includegraphics[width=0.9\linewidth]{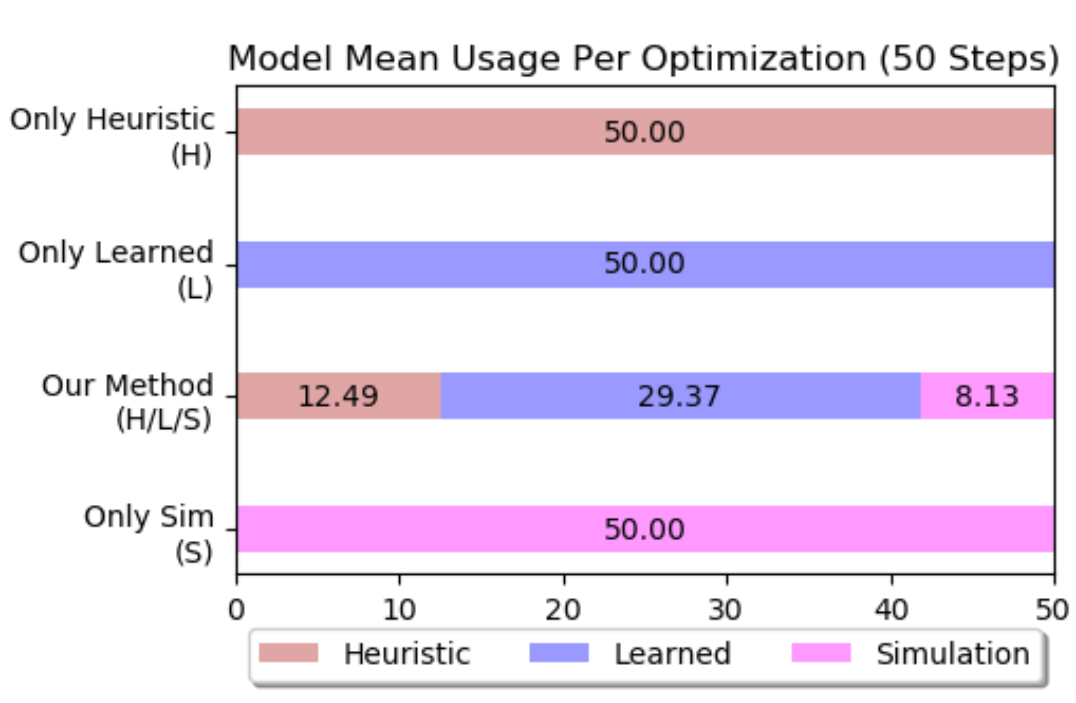}
    \caption{Average usages of the predictive models on Isaac Gym Dataset.}
    \label{fig:usage_sim}
    \vspace{-0.5cm}
\end{figure}

\section{EVALUATIONS}
In our experiments, we investigated the performance of our framework based on two datasets: The \textit{Isaac Gym Dataset} consists of 1000 train and 300 test samples generated in the Isaac Gym simulation environment using a single type of deformable steak. The \textit{Real-World Dataset} consists of 80 train and 20 test samples collected from the real world using six different steaks. We cooked six steaks, four of which were cooked as rare to form deformable object behavior similar to the simulation, and two of them (T-bone steaks) were cooked as well-done to observe rigid object behavior (Fig. \ref{fig:real_world}.left). We introduced the new type of rigid T-bone steaks to investigate if our fine-tuning process can effectively adapt to new out-of-distribution material properties.

\subsection{Performance of the Predictive Models}
We first evaluated the accuracy of the three predictive models used in our framework on both Isaac Gym and Real-world datasets. Table \ref{tab:model_accuracy} shows the heightmap error mean and std values of the three predictive models for both test datasets. 

For the Isaac Gym dataset, the performances of the three models align with their ability to capture complex behaviors. The analytical model performs worse overall compared to the other two predictive models, but is the fastest ($\sim1$ms for a prediction). On the other hand, the simulation model produces the most accurate results since it is the slowest high-fidelity model ($\sim2.5$s for rigid simulation and $\sim5$s for deformable simulation runs). These results clearly show the trade-off between using different quality models in terms of speed and accuracy. 

For the real-world data, we can see that the overall accuracy of all models was reduced significantly. This is expected since the real-world data is more noisy and unpredictable and includes a broader range of steak and material types. Notably, we observe that the learned model suffers the most from the difference between training and test distributions. Since the learned model is trained for a steak that has a specific shape and size from the simulation, the model predicts the output as that steak even though it is given a different one. A model trained on a wider training set may exhibit better performance upon transfer. However, the model learning is not the focus of this paper and we wanted to explore the framework's ability to handle more specialized models. 


\begin{figure*}[t]
    \centering
    \includegraphics[width=\linewidth]{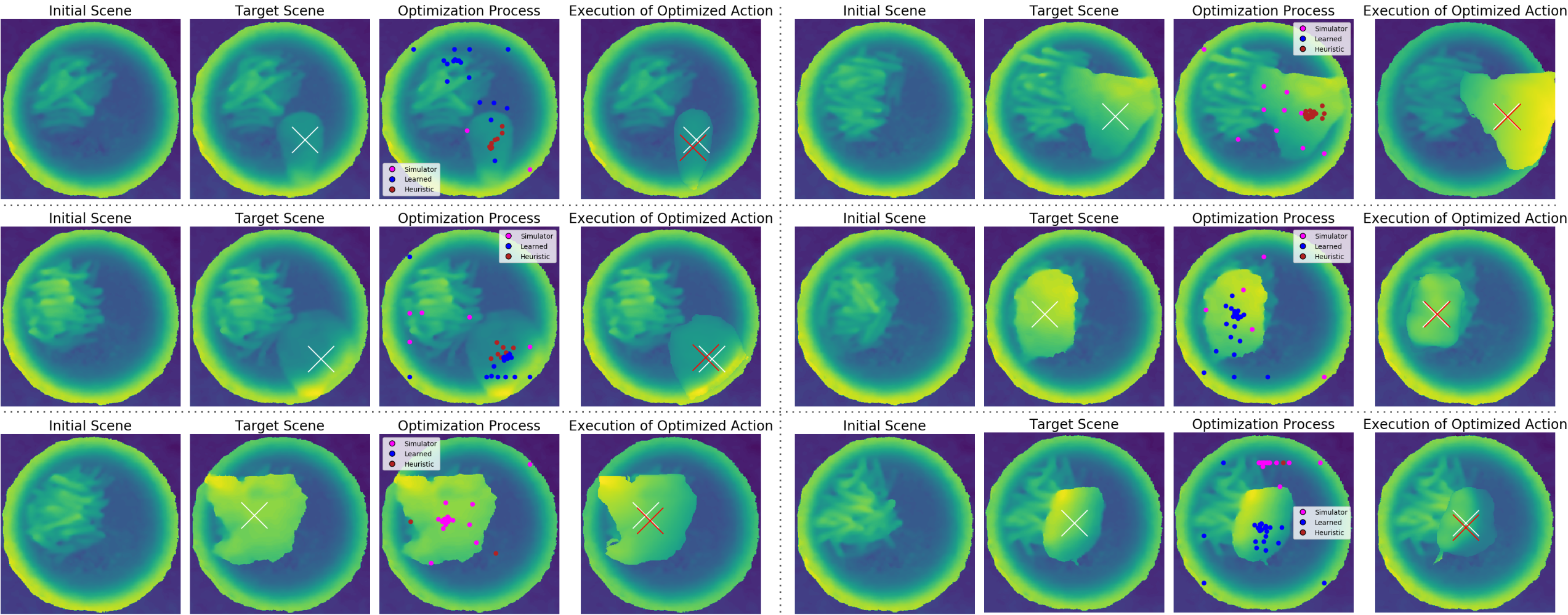}
    \caption{Real-world example results using fine-tuned MDEs for optimizing the plating action based on given initial and target scenes.}
    \label{fig:real_results}
    \vspace{-0.5cm}
\end{figure*}
\begin{figure}[b]
    \centering
    \vspace{-5mm}
    \includegraphics[width=\linewidth]{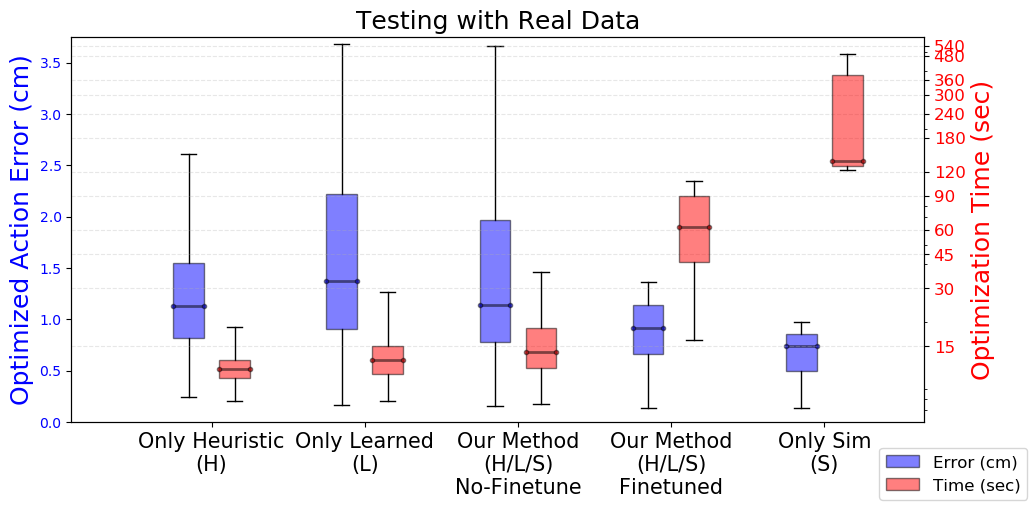}
    \caption{Comparison of our model with the baselines for Real-World dataset. (left axis) optimized action  (right axis, log scale) optimization time.}
    \label{fig:test_real}
\end{figure}
\begin{figure}[b]
    \centering
    \includegraphics[width=0.8\linewidth]{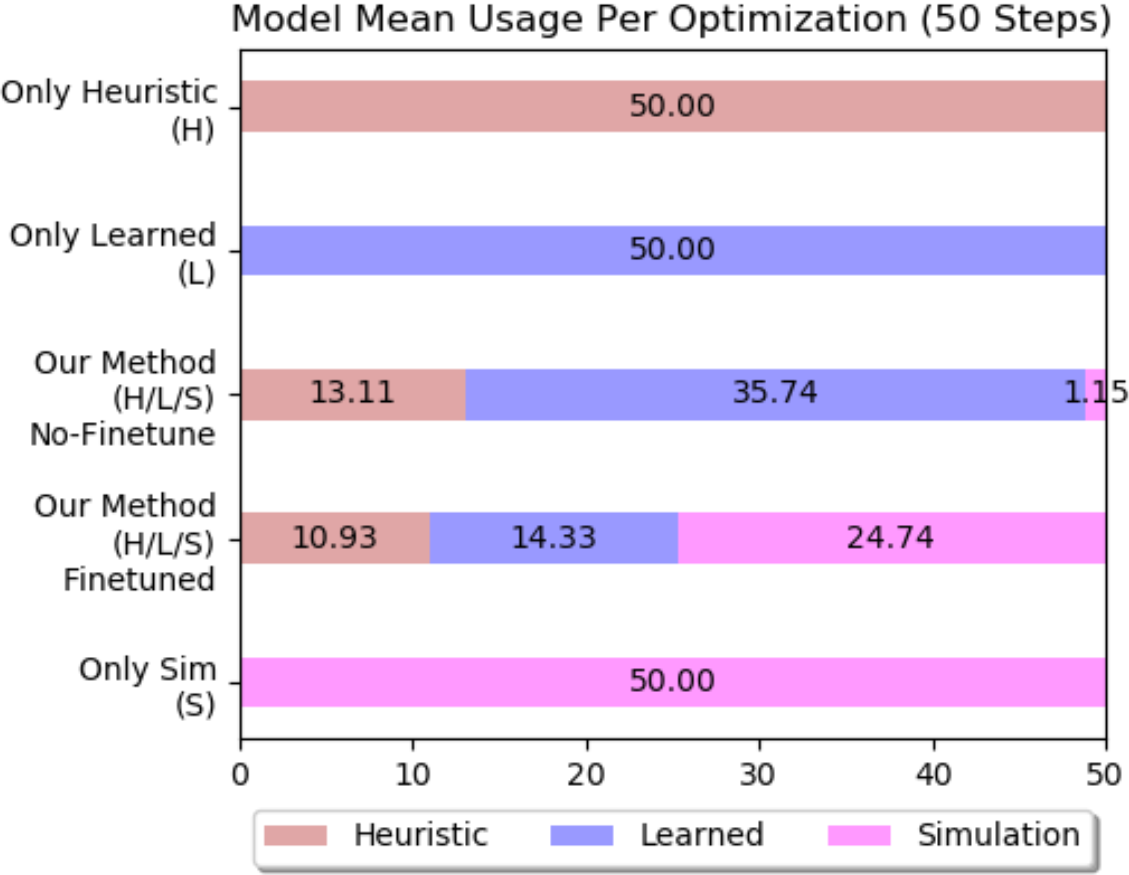}
    \caption{Average usages of the three models on Real-world dataset.}
    \label{fig:usage_real}
\end{figure}
\subsection{Optimization Results with S2S Model Preconditions}
Figure \ref{fig:sim_results} shows example qualitative results of our framework for the plating task with various initial and target scenes. Each example shows four plots from left to right. In the first plot, we can see a heightmap representation of the initial scene and the position of the french fries. The second plot shows the heightmap representation of the target scene and the ground truth of the placing action (white X). In the third plot, the optimization process is illustrated with dots indicating the locations of the action parameters searched by our GP-UCB algorithm. Each dot is colored to show which model was selected to predict the target scene via the MDE model selection algorithm. In the fourth plot, we show the executed scene of the predicted optimal action as a comparison to the target scene shown in the second plot. The red X sign shows the predicted optimal location for the plating action alongside the white ground truth. 

As we can see in the plots, our framework is able to explore the action space successfully before converging to an optimal prediction region. The optimization process also shows that our framework is able to switch between different predictive models according to the initial state and the position of the placing action, thanks to our MDE model selection algorithm. More specifically, we can see that our framework uses the fast analytical model for the flat and empty regions of the plate, where the analytical model mostly aligns with the actual outcomes because the simple planar surface. We also see that our framework is able to switch to other learned and simulation models according to the state-action configuration. The learned model is especially useful for plating locations near the edges or where the steak interacts with the french fries. The model chooses to make predictions with the simulation model for more complex configurations, such as placing locations right on top of the fries or for the gaps between the fries and the plate edges. 

In order to investigate our model's performance on switching between different models and utilizing them according to their accuracy and speed, we compared our model with three other baselines: \textit{Only Heuristic (H)}, \textit{Only Learned (L)}, and \textit{Only Simulation (S)}. Each of these baselines uses only one predictive model, as indicated by their names, during the action optimization phase. \textit{Our Method (H/L/S)} utilizes all three predictive models using S2S MDEs.

Figure \ref{fig:test_sim} shows the comparison of our model with the baseline methods in terms of optimized action error and optimization time. We calculated the optimized action error as the Euclidean distance between the optimal prediction of our model and the actual ground truth action parameters. We can see that \textit{Only Heuristic (H)} and \textit{Only Learned (L)} baselines optimize the actions quickly (less than 15 seconds); however, their optimized action errors are high due to their prediction capabilities. \textit{Only Simulation (S)} model predicts the most accurate actions; however, their optimization time exceeds 240 seconds. \textit{Our Method (H/L/S)} chooses the accurate and fast models using the MDE mechanism and produces accurate action predictions similar to the simulation baseline while keeping the optimization time around 15 seconds.

Figure \ref{fig:usage_sim} demonstrates the average use of each predictive model during the optimization. As expected, the baselines use their corresponding predictive models for all 50 optimization steps. Our method, on the other hand, uses a mix of all models for effective optimization. Combining the insights from our qualitative results with the average model usage numbers, we can see that all of the models are utilized for various cases. Our method mostly prioritizes the usage of the learned model. This is because the learned model can provide sufficiently accurate predictions for most of the actions as it is trained with the simulator data.

\subsection{Sim-to-Real Transfer by Finetuning Pretrained MDEs}
To transfer our pretrained S2S MDEs to real-world S2R MDEs, we used the \textit{Real-World Dataset} that we collected using a Franka robot with a vacuum gripper (Fig \ref{fig:introduction}). We fine-tuned our MDE models by training them using 80 real-world data samples for 500 iterations.

Figure \ref{fig:real_results} shows the qualitative results of our fine-tuned framework, while Figures \ref{fig:test_real} and \ref{fig:usage_real} show the quantitative performance and model distributions, respectively. Our method is able to adapt to real-world observations after the fine-tuning process. As the figures show, our framework utilizes all three predictive models, with more usage of the simulation and less of the learned model after finetuning, and successfully optimizes the action parameters to be near the ground truth values.

In the real world, \textit{Only Heuristic (H)} works better than \textit{Only Learned (L)} because the heuristic model is more adaptive to the different steak shapes. Figure \ref{fig:test_real} shows that the S2S model achieves similar performance to the only learning approach, as the MDE underestimates the sim-to-real deviations for this model. However, by finetuning the S2S into an S2R model, our approach achieves prediction values closer to the only simulation model while still significantly reducing the computation time. 
Table \ref{tab:mde_accuracy}, bottom two rows, shows the accuracy of the MDE models before and after fine-tuning, highlighting the data-efficient adaptation capabilities of our framework.
\subsection{Effect of sample size on Training MDE models}
\begin{figure}[t]
    \centering
    \includegraphics[width=0.9\linewidth]{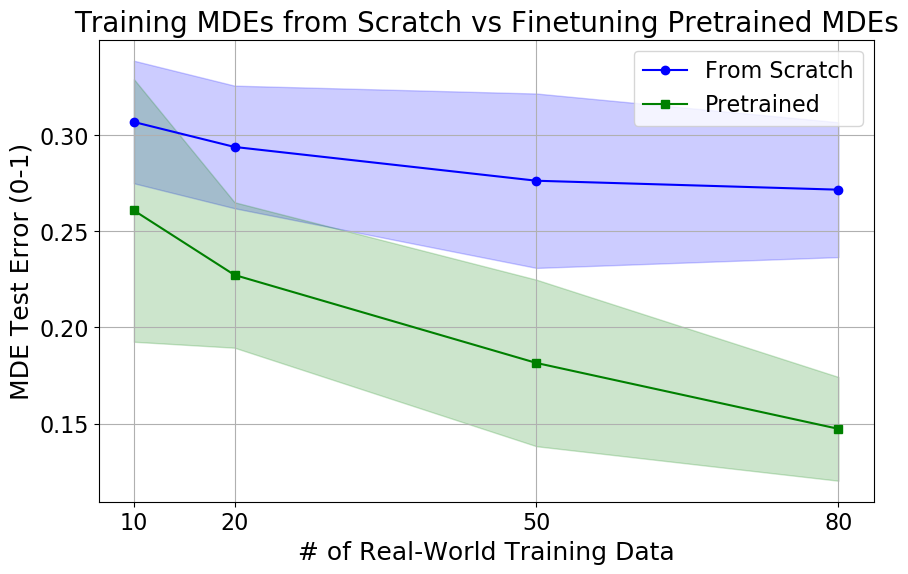}
    \caption{Using pretrained Sim-to-Sim MDEs to fine-tune real-world data vs. training Sim-to-Real MDE from scratch.}
    \label{fig:pretrain}
    \vspace{-0.5cm}
\end{figure}

In this experiment, we investigated the effectiveness of using a pretrained S2S MDE model to fine-tune an S2R MDE on real-world observations. We compared these pretrained MDE models with MDEs that we trained from scratch with only real-world data.  Figure \ref{fig:pretrain} shows the test errors of both types of training approaches for the learned model with increasing numbers of real-world data.  As we can see, the test error of the pretrained MDE decreases significantly as the real-world sample size increases. However, the test error of the from-scratch model fluctuates around the initial value even when the sample size increases. This shows that our method is able to utilize the information it gathered in the pretraining process and smoothly adapt to real-world observations with little data. On the other hand, the from-scratch model quickly overfits the small amount of real-world training data as it is difficult to generalize images with few observations. 
As a consequence, our framework lightens the burden of collecting costly real-world data to learn S2R MDEs and allows quick and effective adaptation to real-world applications.


\section{Conclusion} 
\label{sec:conclusion}

This paper proposes a framework that efficiently optimizes robotic action parameters by integrating multiple predictive models, demonstrated in a robotic food arrangement task. The framework extends the original MDE framework by integrating visual input with material property masks, and uses it as an advanced model selection mechanism for fast and accurate action optimization. This paper also introduces sim-to-sim (S2S) MDEs as one of its main contributions in order to enable faster parameter optimization without require real-world training data. The experiments show that S2S MDEs provide a strong foundation for the efficient transition to sim-to-real (S2R) MDEs through fine-tuning.


\bibliographystyle{IEEEtran}
\bibliography{references}

\addtolength{\textheight}{-12cm}   

%
\end{document}